\documentclass[a4paper,fleqn]{ieeetj}
\usepackage{newtxtext} 
\usepackage{algorithm}
\usepackage{algpseudocode}
\usepackage{graphicx,color}
\usepackage{amsmath}
\usepackage{amssymb}
\usepackage{booktabs}
\usepackage{multirow}

\usepackage[numbers,square,sort&compress]{natbib}
\usepackage{hyperref}
\hypersetup{
    colorlinks=true,
    linkcolor=blue,      
    filecolor=blue,      
    urlcolor=blue,       
    citecolor=blue,      
}

\def\tsc#1{\csdef{#1}{\textsc{\lowercase{#1}}\xspace}}
\tsc{WGM}
\tsc{QE}

\begin{document}
\let\WriteBookmarks\relax
\def\floatpagepagefraction{1}
\def\textpagefraction{.001}


\def\OJlogo{\vspace{-4pt}$<$Society logo(s) and publication title will appear here.$>$}
\def\seclogo{\vspace{10pt}$<$Society logo(s) and publication title will appear here.$>$}

\def\authorrefmark#1{\ensuremath{^{\textbf{#1}}}}

\receiveddate{XX Month, XXXX}
\reviseddate{XX Month, XXXX}
\accepteddate{XX Month, XXXX}
\publisheddate{XX Month, XXXX}
\currentdate{XX Month, XXXX}
\doiinfo{XXXX.2022.1234567}
\title{KDC-Diff: A Latent-Aware Diffusion Model with Knowledge Retention for Memory-Efficient Image Generation.}

\author{Md. Naimur Asif Borno\authorrefmark{1,2}, Md Sakib Hossain Shovon\authorrefmark{3}, Asmaa Soliman Al-Moisheer\authorrefmark{4}, Mohammad Ali Moni\authorrefmark{5,6} }
\affil{Research Assistant, The University of Queensland, 308 Queen St, Brisbane City, QLD 4000, Queensland, Australia.}
\affil{Mechatronics Engineering, Rajshahi University of Engineering \& Technology}
\affil{Researcher, The University of Queensland, 308 Queen St, Brisbane City, QLD 4000, Queensland, Australia.}
\affil{Department of Mathematics and Statistics, Faculty of Science, Imam Mohammad Ibn Saud Islamic University (IMSIU), Riyadh 13318, Saudi Arabia}
\affil{Faculty of Health, Medicine and Behavioural Sciences, The University of Queensland, 308 Queen St, Brisbane City, QLD 4000, Queensland, Australia.}
\affil{AI \& Digital Health Technology Artificial Intelligence and Cyber Futures Institute Charles Sturt University, 308 Queen St, Brisbane City, QLD 4000, Queensland, Australia.}
\corresp{Corresponding author: Mohammad Ali Moni (email: m.moni@uq.edu.au).}
\authornote{Authors:  1808009@student.ruet.ac.bd (Md.N.A. Borno) and sakib.aiub.cs@gmail.com (M.S.H. Shovon) contributed equally.}
\begin{abstract}
The growing adoption of generative AI in real-world applications has exposed a critical bottleneck in the computational demands of diffusion-based text-to-image models. In this work, we propose KDC-Diff, a novel and scalable generative framework designed to significantly reduce computational overhead while maintaining high performance. At its core, KDC-Diff designs a structurally streamlined U-Net with a dual-layered knowledge distillation strategy to transfer semantic and structural representations from a larger teacher model. Moreover, a latent-space replay-based continual learning mechanism is incorporated to ensure stable generative performance across sequential tasks. Evaluated on benchmark datasets, our model demonstrates strong performance across FID, CLIP, KID, and LPIPS metrics while achieving substantial reductions in parameter count, inference time, and FLOPs. KDC-Diff offers a practical, lightweight, and generalizable solution for deploying diffusion models in low-resource environments, making it well-suited for the next generation of intelligent and resource-aware computing systems.

\end{abstract}
\begin{IEEEkeywords}
Stable Diffusion, Continual Learning, Knowledge Distillation, UNet, Text to Image
\end{IEEEkeywords}
\maketitle
\section{Introduction}
In recent years, Text-to-Image (T2I) generative models have garnered widespread attention due to their remarkable capability to transform natural language descriptions into vivid, imaginative visuals, thereby enabling users to effortlessly manifest their creative intent \cite{sawant2021text}. Among these models, diffusion-based approaches have marked a significant leap forward, captivating both academic researchers and creative practitioners with their superior generative performance \cite{zhang2025diff}. T2I methodologies have evolved considerably from the earlier, often inconsistent, GAN-based models to more sophisticated systems incorporating mechanisms like Contrastive Language-Image Pre-training (CLIP), which has substantially enhanced the semantic alignment between text and images \cite{sawant2021text, ding2024clip}. Cutting-edge frameworks such as DALL·E, Imagen, and diffusion-based models leverage advanced architectures to achieve high levels of semantic fidelity and visual accuracy \cite{baldridge2024imagen}.

Among these, Stable Diffusion (SbDf) has emerged as a particularly influential model due to its robustness and generative versatility. However, its iterative denoising process imposes substantial computational overhead \cite{gandikota2023erasing}. Large-scale diffusion models like SbDf have demonstrated exceptional capability in synthesizing photorealistic images that exhibit fine textures, diverse content, and compositional coherence, all while enabling semantic control through textual prompts \cite{kim2025bk}. Beyond image synthesis, SbDf has extended its impact to multi-modal domains such as audio and video generation, further amplifying its relevance in both research and creative sectors \cite{jeong2023power}. Despite these impressive advancements, the computational burden of SbDf remains a formidable challenge. With approximately 890 million parameters, the model demands high memory bandwidth and prolonged inference times due to its iterative noise estimation process, rendering it largely inaccessible to users lacking industrial-grade resources \cite{han2023svdiff, li2023q}. This constraint significantly hinders its applicability in real-time scenarios and practical deployments \cite{zhao2025mixdq}.

Central to the generative success of SbDf is its reliance on the U-Net architecture, which forms the backbone of its image synthesis pipeline \cite{wu2024medsegdiff}. To alleviate the computational strain during inference, several solutions have been proposed, including reducing the number of denoising steps, optimizing architectural efficiency, applying structural pruning, performing quantization, and adopting hardware-aware optimization techniques. While recent innovations have targeted improvements in both efficiency and output quality, many retain the original U-Net design, thereby limiting the extent of achievable enhancements. For example, \cite{sadat2024litevae} proposes an autoencoder-level complexity reduction, whereas \cite{zhao2025mobilediffusion} focuses on accelerating the sampling process. However, both approaches continue to rely on the unmodified U-Net, which remains the primary computational bottleneck. Similarly, \cite{han2023svdiff} introduces parameter-efficient fine-tuning strategies but fails to address U-Net’s core inefficiencies. Although some studies, such as \cite{kim2025bk}, explore lightweight U-Net variants and others like \cite{smith2023continual} propose low-rank adapters for streamlined generation, these strategies fall short in addressing challenges like catastrophic forgetting. Given the pivotal role of U-Net in the diffusion process, its architectural refinement is essential for achieving substantial gains in both generation performance and computational efficiency \cite{wang2024edit}.

While architecturally simplified models often underperform relative to their more complex counterparts, Knowledge Distillation (KD) has emerged as an effective paradigm to bridge this performance gap. In KD, a large, well-trained teacher model imparts its knowledge to a smaller, more efficient student model, thereby enabling the latter to approximate the former’s capabilities while significantly reducing resource demands \cite{nguyen2024swiftbrush}. This approach is especially promising for diffusion-based generative models, where computational overhead is a critical constraint \cite{huang2023knowledge}. By facilitating knowledge transfer, KD methods such as DKDM \cite{xiang2024dkdm}, Progressive KD \cite{gupta2024progressive}, and ADV-KD \cite{mekonnen2024adv} have achieved notable improvements in efficiency. However, a straightforward knowledge transfer between the teacher and the student can lead to suboptimal performance and potential overfitting. Moreover, the application of such distillation strategies to Stable Diffusion models with compact U-Net architectures remains largely unexplored. Therefore, further investigation is warranted to assess their effectiveness in constrained architectural settings.

In parallel, generative models, particularly T2I diffusion systems are notably susceptible to catastrophic forgetting, where acquiring new knowledge disrupts previously learned information. This issue is exacerbated when the model is pruned, as the reduced capacity can impair its ability to retain prior knowledge and maintain performance across diverse tasks.\cite{liang2025diffusion}. Continual Learning (CL) methodologies offer a promising solution to this issue. While replaying previously generated images can mitigate forgetting, generative performance during the early training stages remains a substantial concern \cite{jodelet2023class}. Although low-rank adaptation within cross-attention layers has been explored as a mitigation strategy, its effectiveness diminishes when handling long sequences of evolving concepts \cite{smith2023continual}. Various approaches—including regularization, dynamic architectural adaptation, and generative replay—have been proposed, with experience replay standing out for its promising outcomes \cite{liang2025diffusion}. Nonetheless, the fidelity of synthetic data used in replay remains uncertain, and pixel-space replay methods introduce significant memory and computational burdens during training.

To mitigate the substantial computational overhead associated with Stable Diffusion (SbDf) models, we introduce a novel generative framework that seamlessly integrates a lightweight U-Net backbone, dual-layered knowledge distillation, and latent space-based CL within a unified architecture. At the core of our approach lies a structurally optimized U-Net, meticulously streamlined to retain only the most critical components necessary to maintain image fidelity, thus reducing the overall parameter count to 482 million, significantly lower than conventional SbDf implementations. To address the potential loss in generative capacity due to this architectural compression, we employ a latent space knowledge distillation strategy that fuses soft and hard supervisory signals with intermediate feature-level alignment. This enables the compact student model to effectively inherit semantic and structural knowledge from a larger, high-capacity teacher model. In parallel, to counteract the issue of catastrophic forgetting during continual T2I learning, we propose a latent replay-based CL mechanism that reintroduces internal latent representations of prior tasks, ensuring stable knowledge retention without incurring excessive memory or computational costs. Our key innovation lies in this synergistic integration of architectural efficiency, semantic distillation, and continual adaptation within the diffusion paradigm. Collectively, these contributions enable our model to achieve high-quality image synthesis with significantly reduced resource requirements, advancing the practical deployment of T2I diffusion models in real-world, resource-constrained settings.

\section{Methodology}
\begin{figure*}[!ht]
\centering
\includegraphics[width=1\linewidth]{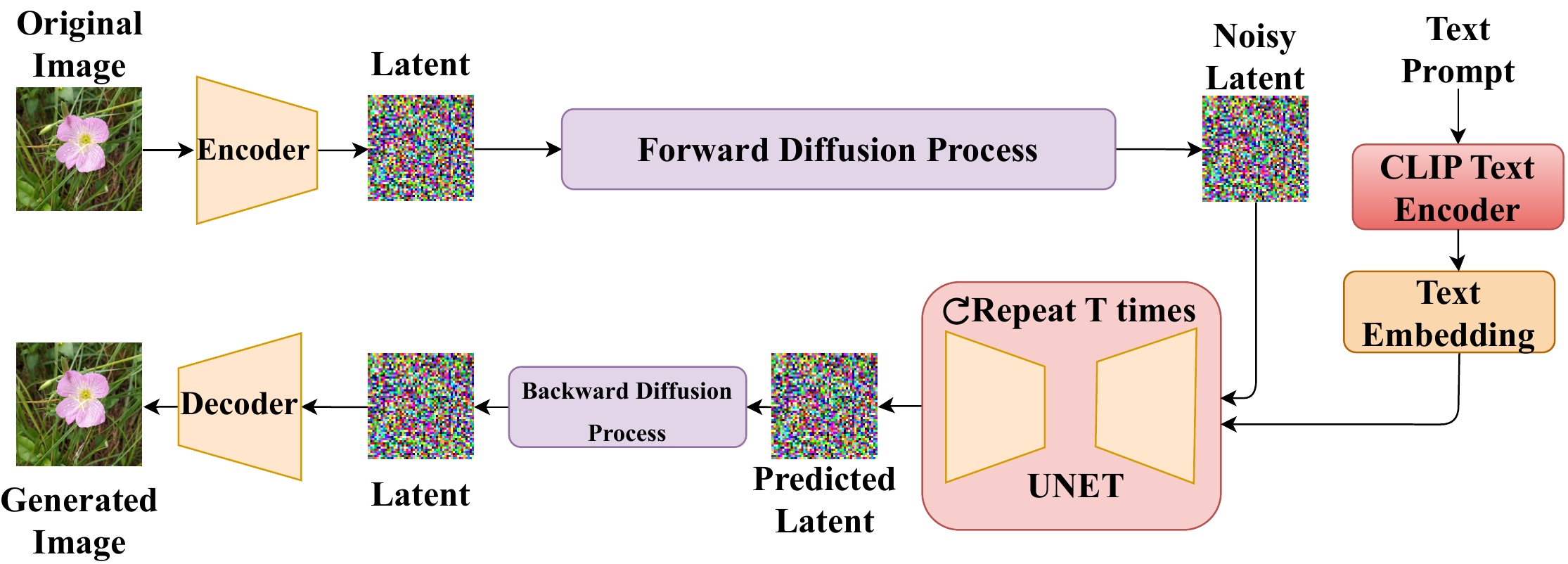}
\caption{Overview of the KDC-Diff framework: (a) illustrates the memory buffer mechanism used to store latent representations of previously learned classes, enabling efficient replay during training on subsequent classes; (b) depicts the overall training architecture, integrating a streamlined U-Net design with CL strategies and knowledge distillation to ensure both efficiency and knowledge retention.}
\label{fig:Figure_1}
\end{figure*}
\subsection{KDC-Diff}
\begin{figure*}[!ht]
\centering
\includegraphics[width=1\linewidth]{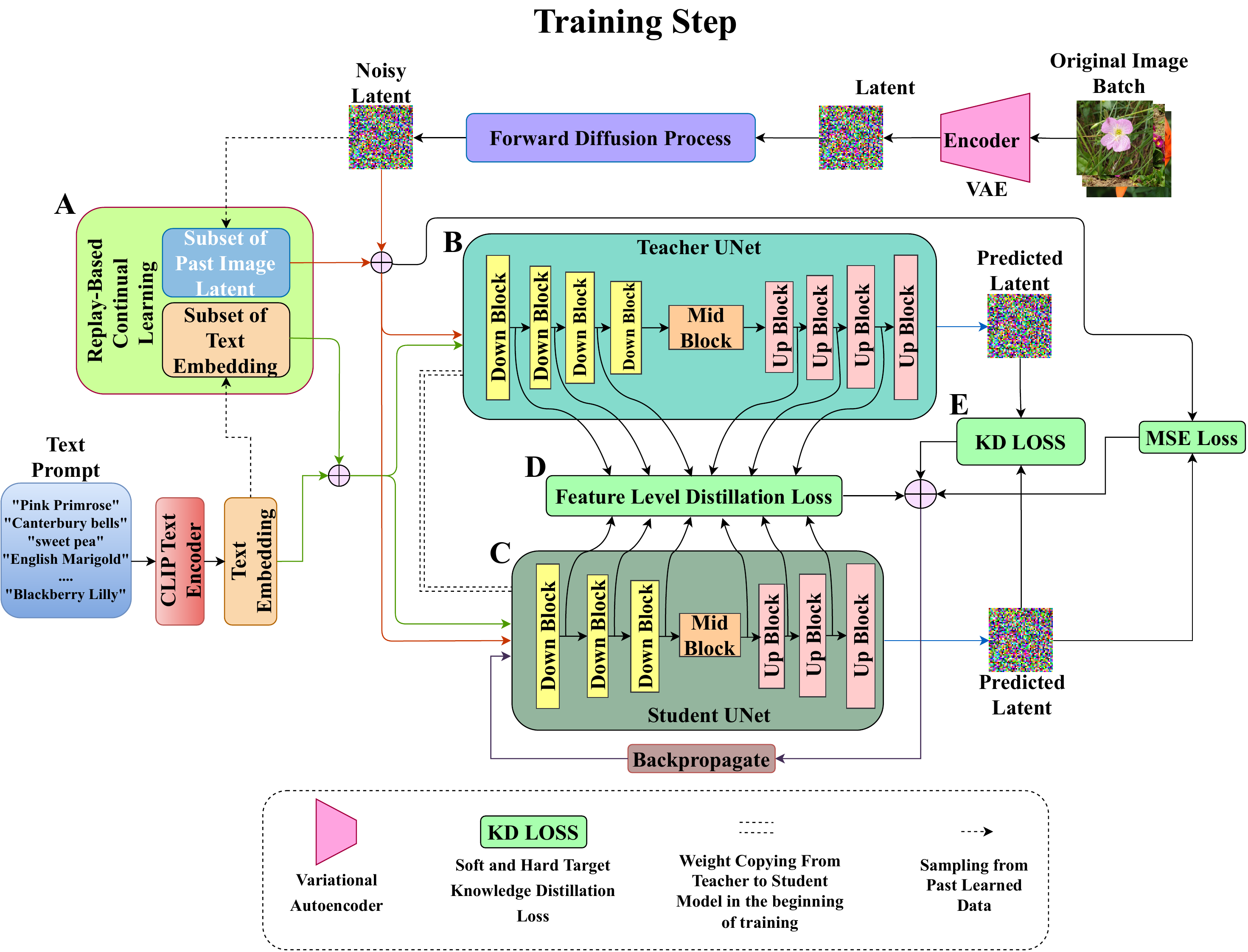}
\caption{\label{fig:Figure_2}Comparison between the baseline U-Net and the optimized KDC-Diff U-Net architecture. Modules outlined in red represent components removed from the teacher model to facilitate a more efficient design, while the highlighted blocks indicate layers eliminated from the original U-Net to derive the lightweight student network.
}
\end{figure*}
To address the growing demand for efficient yet high-performing generative models, we propose a novel architectural framework, KDC-diff, designed to optimize T2I generation under resource constraints. Central to our approach is a streamlined UNet architecture, purposefully restructured to significantly reduce the parameter count while maintaining competitive generative performance. As illustrated in \hyperref[fig:Figure_1]{Fig. \ref*{fig:Figure_1}(b)}, the revised design introduces a lightweight backbone tailored for environments where computational or memory resources are limited.

However, model compression alone often leads to a drop in performance due to the loss of representational capacity. To mitigate this, we incorporate a dual-layered knowledge distillation (KD) framework, which plays a pivotal role in transferring knowledge from a larger, more expressive teacher model \cite{yang2023knowledge}. This framework combines soft and hard target distillation with feature-level alignment, allowing the student model to not only mimic the output distribution of the teacher but also internalize its rich intermediate representations. \hyperref[fig:Figure_1]{Fig. \ref*{fig:Figure_1}(b)}illustrate how this hybrid KD mechanism bridges the gap between accuracy and efficiency, ensuring that the compressed model retains essential semantic understanding.

As generative models are increasingly applied in dynamic and evolving data environments, the need for CL becomes critical. To mitigate catastrophic forgetting, we introduce a novel replay-based CL strategy operating in the latent space. This approach involves selectively storing latent representations instead of pixel representations in a memory buffer and revisiting them during training, thereby promoting stable performance across sequential tasks \cite{maracani2021recall}. The latent storage mechanism is depicted in \hyperref[fig:Figure_1]{Fig. \ref*{fig:Figure_1}(a)}, while its integration into the overall training framework is shown in \hyperref[fig:Figure_1]{Fig. \ref*{fig:Figure_1}(b)}.

The following subsections provide an in-depth discussion of these components, detailing the architectural innovations in UNet design, the methodologies underpinning the KD framework, and the implementation of CL strategies. These enhancements collectively contribute to a lightweight yet highly effective model designed for next-generation generative AI tasks in a resource-constrained environment.
\subsubsection{UNet Modification}

\hyperref[fig:Figure_2]{Fig. \ref*{fig:Figure_2}} illustrates the detailed architectures of both the Teacher UNet and the Student UNet, as introduced in \hyperref[fig:Figure_1]{Fig. \ref*{fig:Figure_1}(b)}. To initiate the optimization process, we conducted a comprehensive evaluation of ten competitive SbDf models to determine the most effective baseline architecture. Among these, the Realistic Vision V51 model was selected due to its superior performance across four key evaluation metrics, as summarized in  \hyperref[table:Table_1]{Table \ref*{table:Table_1}}. In addition to its strong generative capability, it demonstrated favorable computational efficiency, with an inference time of 11.8344 seconds, as reported in  \hyperref[table:Table_2]{Table \ref*{table:Table_2}}. This architecture, referred to as the Original UNet, served as the foundation for our subsequent modifications.

The Original UNet comprises three CrossAttnDownBlock2D layers and one DownBlock2D layer in the encoder, a central UNetMidBlock2DCrossAttn layer, and three CrossAttnUpBlock2D layers along with one UpBlock2D layer in the decoder. Each CrossAttnDownBlock2D and CrossAttnUpBlock2D contains three R-T block pairs, where “R” denotes a ResNet block \cite{he2016deep} and “T” represents a Transformer2DModel block \cite{jaegle2021perceiver}. The Transformer2DModel block incorporates cross-attention mechanisms, enabling the model to capture long-range dependencies, which is critical for generating semantically coherent and visually rich outputs.

The central aim of our study is to design a lightweight UNet architecture that matches the performance of the original model while significantly improving computational efficiency. Motivated by prior work on model compression \cite{sanh2019distilbert}, which suggests that reducing network depth can preserve performance while enhancing speed, we first explored the effect of reducing the number of R-T block pairs in each attention module. Specifically, we removed one R-T block from each CrossAttnDownBlock2D and CrossAttnUpBlock2D, effectively lowering the model’s depth and parameter count.

Building on this, we further investigated the utility of specific components within the encoder-decoder pathway. Studies such as \cite{kim2023bk, kim2022cut} have shown that the innermost layers of UNet architectures often contain redundant filters that can be pruned with negligible impact on performance. Guided by this insight, we experimented with the removal of the DownBlock2D layer. However, this led to a noticeable degradation in image quality, as the generated outputs deviated significantly from those produced by the original model. Interestingly, when both the DownBlock2D and UpBlock2D layers were simultaneously removed, the quality of the generated images substantially improved, closely matching the fidelity of the original outputs. This result suggests that while these individual components may contribute modestly, their joint removal may help avoid overfitting or redundancy, leading to a more efficient and balanced architecture. To ensure that the modified UNet performs similarly to the teacher model, we transferred the weights from the teacher model to the student model. Therefore, these layers were completely removed from the modified U-Net architecture to enhance efficiency. The modified Unet is also illustrated in \hyperref[fig:Figure_2]{Fig. \ref*{fig:Figure_2}}.

\subsubsection{Knowlege Distillation}
KD, a paradigm of model compression, involves training a more compact student model to emulate the behaviour of a larger and more intricate teacher model \cite{beyer2022knowledge}. To enable our lightweight diffusion model to achieve performance comparable to a larger, more expressive teacher, we design a targeted KD framework tailored for generative tasks. The primary goal is to transfer both high-level predictive behavior and internal semantic representations from the teacher to the student, while keeping the resource footprint minimal, a necessity in real-world deployment scenarios.

Unlike conventional KD techniques that rely heavily on pixel-level reconstruction or high-resolution logits, our approach performs distillation in the latent space, leveraging compact yet semantically rich representations. This allows the student model to efficiently acquire the essential knowledge encoded by the teacher without incurring significant computational or memory overhead. Soft target distillation, a pivotal component of this framework, focuses on aligning the softened probabilistic outputs of the student model with those of the teacher \cite{li2022soft}. The loss associated with this alignment is computed using the Kullback-Leibler (KL) divergence, which quantifies the discrepancy between the probability distributions produced by the two models, facilitating a seamless transfer of nuanced predictive patterns from the teacher to the student \cite{hwang2023comparison}. The KL divergence can be computed using,
\begin{equation}
    \mathcal{L}_{\text{soft}} = T^2 \cdot \frac{1}{N} \sum_{i=1}^{N} q_{i}^T \left[ \log(q_{i}^T) - \log(q_{i}^S) \right]
\end{equation}
here, $L_{soft}$ is the soft target loss, $T$ is the temperature for softening the logits, $N$ is the number of samples in the batch, $q_{i,j}^T$ and $q_{i,j}^S$ are the softened probabilities for the 
i-th sample from the teacher and student models, respectively.

To further reinforce alignment with the teacher's decision boundaries, we also employ hard target distillation which is computed using the cross-entropy between the outputs of the teacher and student models, ensuring alignment with the true target distribution using \cite{guo2024shared}, 
\begin{equation}
    \mathcal{L}_{\text{hard}} = \frac{1}{N} \sum_{i=1}^{N} \left( -y_i \log(q_{i}^S) - y_i \log(q_{i}^T) \right)
\end{equation}
$L_{hard}$ is the hard target loss, $N$ is the number of samples in the batch, $y_i$ is the true label for i-th sample, $q_{i}^S$ and $q_{i}^T$ are the predicted probabilities from the student and teacher models, respectively, conditioned on the same input.
In addition to output alignment, we introduce a feature-level distillation mechanism that operates in the latent space of the diffusion pipeline. This enables the student model to assimilate semantic understanding embedded in the intermediate layers of the teacher, enhancing output quality, generalization, and spatial-contextual coherence. Rather than relying on early feature maps or full-resolution comparisons, our method supervises the student using compact latent representations. These are extracted from and aligned with corresponding teacher representations, computed as:
\begin{equation}
    \mathcal{L}_{\text{feature}} = \frac{1}{N} \sum_{i=1}^{N} \| \Phi_T(x_i) - \Phi_S(x_i) \|_2^2
\end{equation}
here, $L_{feature}$ is the feature-based distillation loss, $\Phi_T(x_i)$ is the latent feature representation from the teacher model for the i-th input $x_i$, $\Phi_S(x_i)$ is the latent feature representation from the student model for the i-th input $x_i$. This integration of latent-space feature distillation represents a key contribution of our work, enabling efficient and resource-aware transfer of generative capabilities without relying on pixel-wise reconstruction or full-image storage.

\subsubsection{Latent Space Replay Based CL}
\begin{figure*}[!ht]
    \centering
    \includegraphics[width=.7\linewidth]{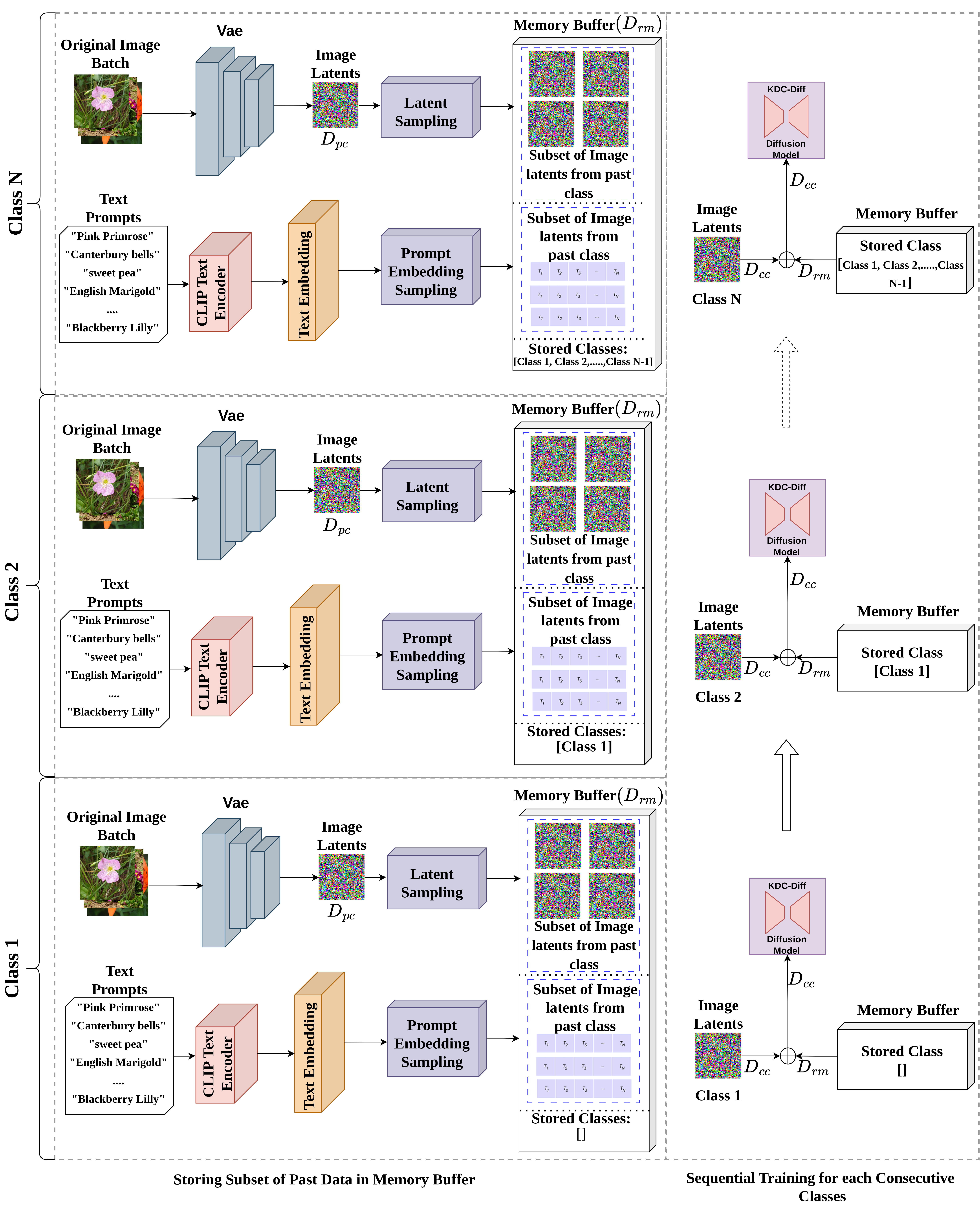}
    \caption{\label{fig:Figure_3}Illustration of sequential training in a diffusion model using latent-space replay for continual learning. Latent representations from multiple classes are stored in a fixed-size buffer and replayed during training on new tasks. This mitigates catastrophic forgetting by preserving class diversity, enabling efficient memory use and stable generation quality across tasks. }
    \label{fig:Figure_3}
\end{figure*}
\begin{table*}[!ht]
\centering
\caption{Comparison of various diffusion models (including KD-based) across two datasets: Oxford-102-Flower and Butterfly \& Moths 100 Species. Metrics include FID, CLIP Score, KID, and LPIPS. Lower FID, KID, and LPIPS are better. To clarify knowledge distillation models, we define S as student as T as Teacher.}
\begin{tabular}{c|lcccc}
\toprule
\textbf{Dataset} & \textbf{Model} & \textbf{FID ($\downarrow$)} & \textbf{CLIP ($\uparrow$)} & \textbf{KID ($\downarrow$)} & \textbf{LPIPS ($\downarrow$)} \\
\midrule
\multirow{22}{*}{\rotatebox[origin=c]{90}{Oxford-102-Flower}} 
&stable-diffusion-v1-4   & 285.00   & 28.00    & 0.0043 ± 0.0700 & 0.67 \\ 
&stable-diffusion-v1-5   & 259.48   & 28.5031  & 0.0096 ± 0.0916 & 0.66  \\ 
&stable-diffusion-2      & 308.05   & 30.5607  & 0.0067 ± 0.0825 & 0.71  \\ 
&stable-diffusion-2-base & 266.36   & 30.8017  & 0.0059 ± 0.0916 & 0.65  \\ 
&stable-diffusion-2-1    & 277.94   & 28.92    & 0.0051 ± 0.0830 & 0.69  \\ 
&anything-v5             & 302.46   & 29.04    & 0.0085 ± 0.1095 & 0.73  \\ 
&realistic-vision-v51    & 251.15  & 28.96    & 0.0040 ± 0.0387 & 0.64 \\ 
&anything-midjourney     & 255.73   & 27.85    & 0.0078 ± 0.0900 & 0.65  \\ 
&urpm-v13                & 269.50   & 28.94    & 0.0097 ± 0.1044 & 0.65  \\ 
&hassaku-v13             & 273.40   & 28.01    & 0.0042 ± 0.0755 & 0.70  \\ 
& S: anything-v5 , T: realistic-vision-v51 & 235.00 & 24.4768 & $0.0096 \pm 0.0648$ & 0.66 \\
& S: stable-diffusion-v1-5, T: realistic-vision-v51 & 247.23 & 23.06 & $0.0101 \pm 0.0625$ & 0.70 \\
& S: hassaku-v13, T: realistic-vision-v51 & 241.41 & 24.6603 & $0.0007 \pm 0.0700$ & 0.67 \\
& S: anything-mid-journey, T: realistic-vision-v51 & 219.81 & 23.20 & $0.0110 \pm 0.0632$ & 0.66 \\
& S: realistic-vision-v51, T: anything-mid-journey & 244.24 & 28.54 & $0.0032 \pm 0.0818$ & 0.63 \\
& S: anything-mid-journey, T: anything-mid-journey & 237.20 & 21.45 & $0.0079 \pm 0.0990$ & 0.64 \\
& S: urpm-v13, T: anything-mid-journey & 256.22 & 28.00 & $0.0091 \pm 0.0510$ & 0.65 \\
& S: anything-v5, T: anything-mid-journey & 230.87 & 25.60 & $0.0091 \pm 0.0910$ & 0.64 \\
& S: KDC-diff-x, T: anything-mid-journey & 255.00 & 24.00 & $\textbf{0.0002} \pm \textbf{0.0510}$ & 0.65 \\
& S: KDC-diff-CL, T: realistic-vision-v51 & 280.00 & 28.00 & $0.0066 \pm 0.0963$ & 0.85 \\
& S: KDC-diff-y, T: anything-v5 & 230.80 & 23.00 & $0.0117 \pm 0.1025$ & 0.67 \\
& \textbf{KDC-diff (Proposed)} & \textbf{177.3690} & \textbf{28.733} & $0.0043 \pm 0.0912$ & \textbf{0.62} \\
\midrule
\multirow{11}{*}{\rotatebox[origin=c]{90}{Butterfly \& Moths}} 
& stable-diffusion-v1-5 & 332.56 & 30.12 & $0.0047 \pm 0.0616$ & 0.653 \\
& stable-diffusion-2 & 339.69 & 30.28 & $0.0037 \pm 0.0669$ & 0.672 \\
& stable-diffusion-2-base & 323.79 & 31.82 & $0.0038 \pm 0.0693$ & 0.644 \\
& stable-diffusion-2-1 & 344.25 & 28.42 & $0.0030 \pm 0.0581$ & 0.708 \\
& anything-v5 & 345.71 & 27.35 & $0.0022 \pm 0.0648$ & 0.749 \\
& stable-diffusion-v1-4 & 356.78 & 25.90 & $0.0052 \pm 0.0889$ & 0.794 \\
& realistic-vision-v51 & 302.58 & 28.04 & $0.0020 \pm 0.0509$ & 0.640 \\
& urpm-v13 & 343.92 & 28.97 & $0.0025 \pm 0.0540$ & 0.659 \\
& anything-mid-journey & 343.76 & 28.56 & $0.0040 \pm 0.0566$ & 0.650 \\
& hassaku-v13 & 339.73 & 28.09 & $0.0036 \pm 0.0565$ & 0.655 \\
& \textbf{KDC-diff (Proposed)} & \textbf{297.66} & \textbf{33.89} & $\textbf{0.0017} \pm \textbf{0.0435}$ & \textbf{0.581} \\
\bottomrule
\end{tabular}
\label{table:Table_1}
\end{table*}
Replay-based CL  offers an effective solution to the problem of catastrophic forgetting, which is particularly pronounced in diffusion models trained incrementally on evolving datasets \cite{bagus2021investigation}. Traditional replay strategies typically store and reuse pixel-level images from previous tasks during training. Although this helps to retain past knowledge, it incurs substantial memory and computational costs, especially in high-resolution generative models. 

To address these limitations, we adopt a more efficient approach by storing latent space representations instead of raw images. These latent vectors, extracted from the encoder of the diffusion model, retain essential semantic information while significantly reducing memory overhead. As the model begins training on data from a new class, the replay memory is augmented with data from the current class. As depicted in \hyperref[fig:Figure_3]{Fig. \ref*{fig:Figure_3}}, this study incorporates a replay-based CL strategy to mitigate the issue of catastrophic forgetting effectively. In scenarios encompassing $N$ classes, the model undergoes initial training on the first class, with a selectively curated subset of its data stored in the replay memory, formalized as,
\begin{equation}
    D_{rm}=D_{rm}+\hat f(D_{pc})
\end{equation}
here $\hat f$ is the function that samples data from previous class data, $D_{pc}$ is the entire data of previous class, $D_{rm}$ is the memory buffer containing previously learned classes. This stored subset is then integrated with the data of the second class during its training phase, and the process is iteratively extended to subsequent classes, ensuring that representative subsets from each class are systematically preserved within the replay memory depicted as,
\begin{equation}
    D_{cc}=D_{cc}+D_{rm}
\end{equation}
Here, $D_{cc}$ is the current class data on which the model is trained. This approach enables the model to maintain a delicate equilibrium between retaining prior knowledge and assimilating new information. By reintroducing critical data from previously encountered tasks into subsequent training cycles, the methodology ensures that the model adapts effectively to evolving datasets while minimizing the adverse effects of catastrophic forgetting.

\subsubsection{Loss Function}
We introduce three hyperparameters, namely $\alpha$, $\beta$, and $\gamma$, to effectively regulate the contribution of each distinct distillation loss within our framework. To train the model, the overall loss function is defined as follows:
\begin{equation}
    L_{total}=\alpha L_{soft}+(1-\alpha) L_{hard}+\beta L_{feature}+ \gamma L_{mse}
\end{equation}
Here, the $L_{soft}$ is the soft target loss, $L_{hard}$ is the hard target, and $L_{feature}$ is the feature level distillation loss as depicted in (4), (5), and (6). The $L_{mse}$ is the mean squared error loss of the output of the model and the original data. This computed loss propagates backward through the network during backpropagation to update the model's parameters.
\subsubsection{Evaluation Metrics}
The model's efficiency was appraised utilizing four distinct evaluation metrics: Frechet Inception Distance (FID), CLIP score, Kernel Inception Distance (KID), and Learned Perceptual Image Patch Similarity (LPIPS). FID quantifies the statistical divergence between the feature distributions of authentic and generated images, offering a measure of their relative quality \cite{nunn2021compound}. The governing equation can be expressed as,
\begin{equation}
FID(Re, Ge) = \|\mu_{Re} - \mu_{Ge}\|_2^2 + \operatorname{Tr}(\Sigma_{Re} + \Sigma_{Ge} - 2\sqrt{\Sigma_{Re} \Sigma_{Ge}})
\end{equation}
here, $\mu_{Re}, \Sigma_{Re}$ the mean vector and covariance matrices of the features for the real images, $\mu_{Ge}$, $\Sigma_{Ge}$ are the mean vector and covariance matrices of the features for the generated images. $Tr$ is the trace operator, which sums the diagonal elements of a matrix.\\
KID, conversely, calculates the squared Maximum Mean Discrepancy (MMD) between the feature spaces of real and synthetic images, providing a robust comparison of their underlying distributions \cite{horak2021topology}. For calculating KID, the following equation is used,
\begin{equation}
    KID=MMD(P_{Og},P_{Gn})^2
\end{equation}
here, $P_{Og}$ is the original image and $P_{Gn}$ is the generated image.\\
The CLIP score gauges the semantic congruence between an image and its counterpart, evaluating how visual content aligns with the prescribed textual description \cite{wang2023exploring}. The calculating equation,
\begin{equation}
    CLIP=w*max(cos(a,b),0)
\end{equation}
Here, $a$ and $b$ are the visual and textual embedding, $cos(a,b)$ computes the cosine similarity between two entities, and $w$ is the scaling factor.

Meanwhile, LPIPS assesses perceptual similarity by leveraging deep convolutional networks pre-trained on image classification tasks, thereby capturing both low-level details and high-level semantic structures to offer a human-like evaluation of visual fidelity \cite{cheng2021perceptual}. Furthermore, to assess the computational complexity of the models, we employed Floating Point Operations (FLOPs), also known as Multiply-Accumulate (MAC) operations—which are widely recognized as a standard metric for measuring computational cost.


\section{Results and Analysis}
\begin{table*}[!ht]
\centering
\caption{Inference time, FLOPs, and total parameter count of various diffusion models on Oxford-102-Flower and Butterfly \& Moths 100 Species datasets. To clarify knowledge distillation models, we define S as student as T as Teacher.}
\begin{tabular}{c|lccc}
\toprule
\textbf{Dataset} & \textbf{Model} & \textbf{Inference Time (s)} & \textbf{Total Parameters} & \textbf{FLOPs (GMac)} \\
\midrule
\multirow{22}{*}{\rotatebox[origin=c]{90}{Oxford-102-Flower}} 
&stable-diffusion-v1-4   & 12.0310  & 859,520,964  & 339.01 \\ 
&stable-diffusion-v1-5   &  12.9523  & 859,520,964  & 339.01 \\ 
&stable-diffusion-2      &  30.7353  & 865,910,724  & 339.50 \\ 
&stable-diffusion-2-base &  12.6776  & 865,910,724  & 339.50 \\ 
&stable-diffusion-2-1    &  13.0200  & 865,910,724  & 339.50 \\ 
&anything-v5             &  12.6479  & 859,520,964  & 339.01 \\ 
&realistic-vision-v51    &  11.8344  & 859,520,964  & 339.01 \\ 
&anything-midjourney     &  12.1890  & 859,520,964  & 339.01 \\ 
&urpm-v13                &  12.3175  & 859,520,964  & 339.01 \\ 
&hassaku-v13             &  12.1510  & 859,520,964  & 339.01 \\ 
& S: anything-v5, T: realistic-vision-v51 & 12.7325 & 859,520,964 & 339.01 \\
& S: stable-diffusion-v1-5, T: realistic-vision-v51 & 12.1825 & 859,520,964 & 339.01 \\
& S: hassaku-v13, T: realistic-vision-v51 & 12.5410 & 859,520,964 & 339.01 \\
& S: anything-mid-journey, T: realistic-vision-v51 & 12.1684 & 859,520,964 & 339.01 \\
& S: realistic-vision-v51, T: anything-mid-journey & 12.6943 & 859,520,964 & 339.01 \\
& S: anything-mid-journey, T: anything-mid-journey & 13.0298 & 859,520,964 & 339.01 \\
& S: urpm-v13, T: anything-mid-journey & 11.7840 & 859,520,964 & 339.01 \\
& S: anything-v5, T: anything-mid-journey & 12.0915 & 859,520,964 & 339.01 \\
& S: KDC-diff-x, T: anything-mid-journey & 12.7688 & 482,346,884 & 228.85 \\
& S: KDC-diff-CL, T: realistic-vision-v51 & 13.0230 & 482,346,884 & 228.85 \\
& S: KDC-diff-y, T: anything-v5 & 12.3980 & 482,346,884 & 228.85 \\
& \textbf{KDC-diff (Proposed)} & \textbf{7.8540} & \textbf{482,346,884} & \textbf{228.85} \\
\midrule
\multirow{11}{*}{\rotatebox[origin=c]{90}{Butterfly \& Moths}} 
& stable-diffusion-v1-5 & 12.22 & 859,520,964 & 339.01 \\
& stable-diffusion-2 & 31.01 & 865,910,724 & 339.50 \\
& stable-diffusion-2-base & 11.55 & 865,910,724 & 339.50 \\
& stable-diffusion-2-1 & 30.94 & 865,910,724 & 339.50 \\
& anything-v5 & 12.22 & 859,520,964 & 339.01 \\
& stable-diffusion-v1-4 & 12.68 & 859,520,964 & 339.01 \\
& realistic-vision-v51 & 12.17 & 859,520,964 & 339.01 \\
& urpm-v13 & 12.34 & 859,520,964 & 339.01 \\
& anything-mid-journey & 12.23 & 859,520,964 & 339.01 \\
& hassaku-v13 & 12.25 & 859,520,964 & 339.01 \\
& \textbf{KDC-diff (Proposed)} & \textbf{7.94} & \textbf{482,346,884} & \textbf{228.85} \\
\bottomrule
\end{tabular}
\label{table:Table_2}
\end{table*}
\begin{figure*}[!ht]
    \centering
    \includegraphics[width=0.75\linewidth]{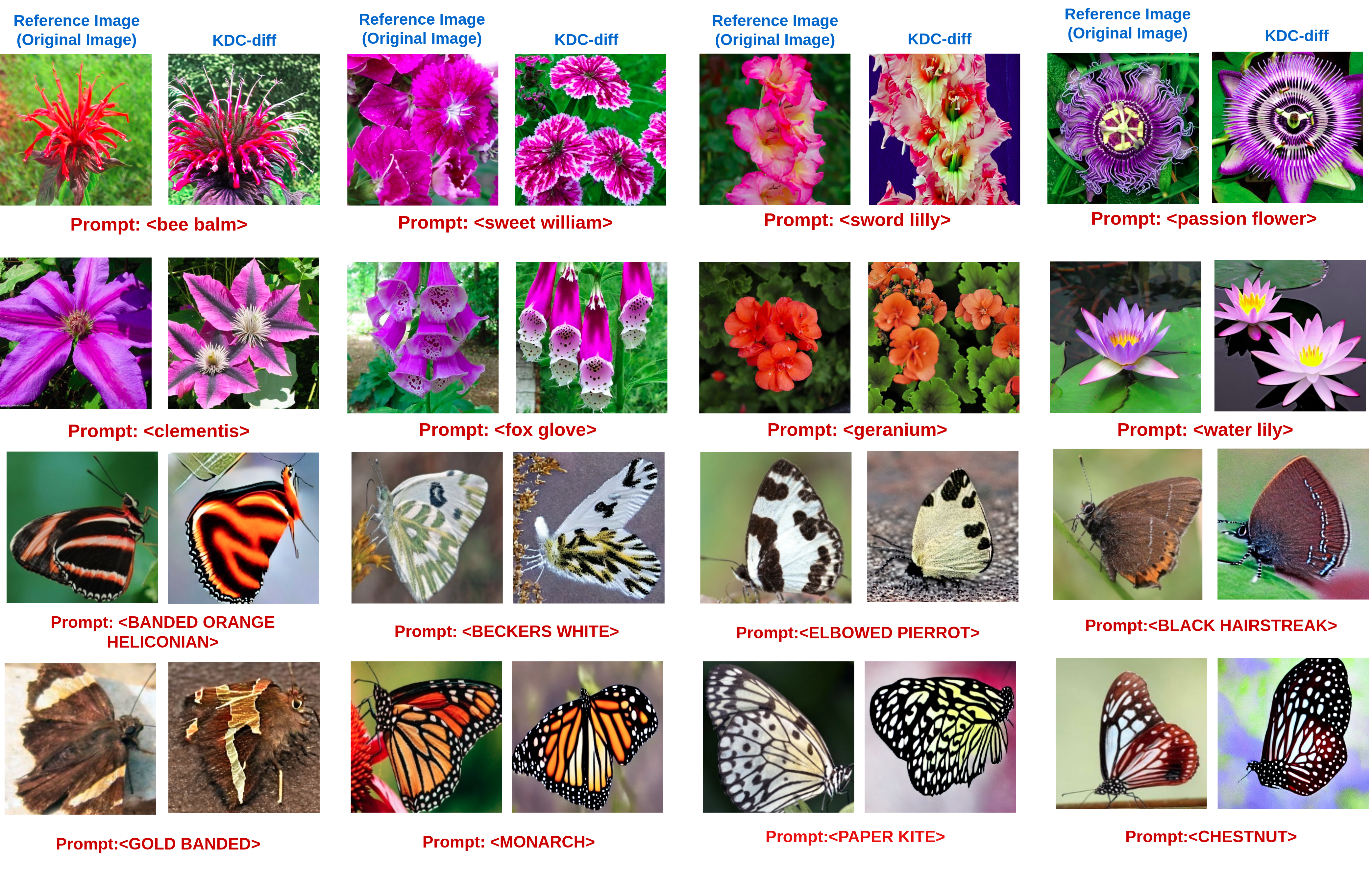}
    \caption{\label{fig:Figure_4}Visual Comparison of Original and Generated Images for Different Classes in Both Datasets. Our proposed methods produced high-quality, color-enhanced images from scratch under computational constraints.}
\end{figure*}
We evaluated KDC-diff on two datasets: the Oxford 102 Flower Dataset, containing 6,149 training images and 2,040 test images at 512×512 resolution, and the Butterfly \& Moth 100 Species Dataset, which comprises 12,594 training images to leverage robustness and generalization capability. Training was uniformly limited to all models, including baselines, with a batch size of 1 optimized for VRAM constraints and a learning rate of \(5 \times 10^{-5}\). Gradient checkpoint and mixed precision training were used to maximize memory efficiency. For fairness, all baseline models were retrained from scratch under identical conditions, eliminating hardware or configuration bias. Using standardized prompts and seeds, evaluation metrics (FID, CLIP, KID, and LPIPS), inference time and FLOPs count were measured on the same T4 GPU.

We devised several KD model configurations to benchmark our model against various SbDf models. We evaluated their performance on the Oxford 102 Flower Dataset using a suite of evaluation metrics, including FID, CLIP, KID, and LPIPS. The comprehensive results are presented in  \hyperref[table:Table_1]{Table \ref*{table:Table_1}}.
\begin{figure*}[!ht]
    \centering
    \includegraphics[width=0.65\linewidth]{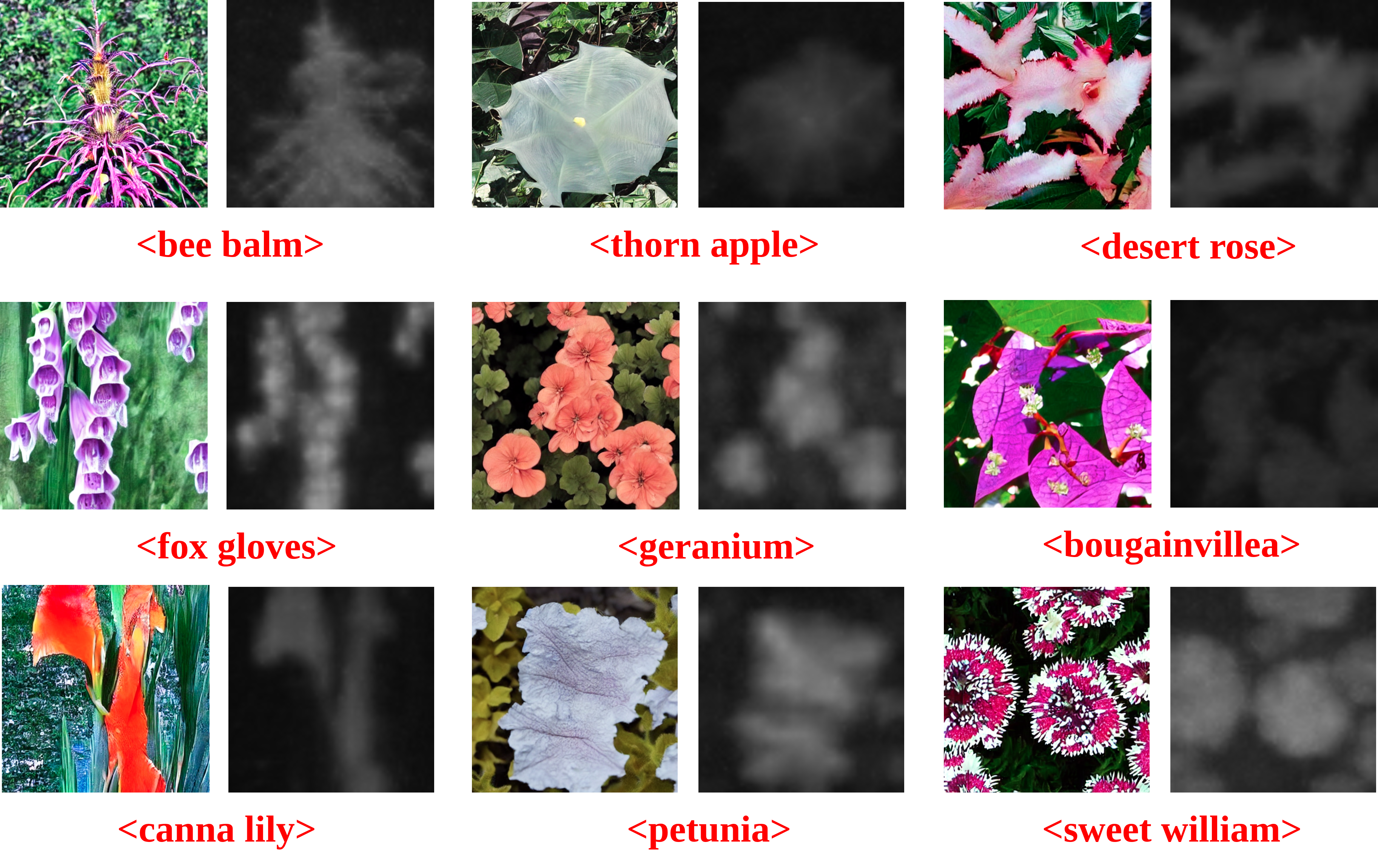}
    \caption{\label{fig:Figure_5} Attention map visualization during image generation. The grayscale heatmap highlights spatial regions with higher attention weights, indicating where the model focused most while generating the corresponding image. These heatmaps captured the semantic structure so well from the reference images.}
\end{figure*}

Our proposed UNet architecture was integrated with various teacher models to facilitate KD. One such configuration, denoted as KDC-diff-x, employed the teacher model Anything-Midjourney. While this combination exhibited an exceptional KID score of 0.0002±0.0510, it underperformed on other metrics compared to alternative models. Furthermore, removing CL from this configuration terms as KDC-diff-CL led to a significant decline in overall performance, as evidenced by an inflated FID score of 280—higher than any other model listed in  \hyperref[table:Table_1]{Table \ref*{table:Table_1}}.

In an alternative configuration, the KDC-Diff UNet was paired with the teacher model Anything V5, resulting in the variant KDC-Diff-y. While this setup achieved a respectable FID score of 230.8 and a KID score of 0.0117±0.1025, its performance on other key metrics—specifically a CLIP score of 23 and an LPIPS score of 0.67—was suboptimal, ranking among the lowest in the evaluation.

In contrast, our proposed model, which integrates the optimized lightweight UNet with CL and is guided by the Realistic Vision v51 teacher, demonstrates superior performance across most evaluation metrics. This configuration achieves the lowest FID score of 177.37, the highest CLIP score of 28.73, and an LPIPS score of 0.62, highlighting its strong visual and semantic fidelity. While its KID score is slightly higher than that of KDC-Diff-x, the model’s consistent excellence across the remaining metrics underscores its robustness, even in resource-constrained settings.

To evaluate the robustness of the proposed model, its performance was rigorously compared against the aforementioned SbDf models using the Butterfly \& Moth 100 Species Dataset, which is also given in  \hyperref[table:Table_1]{Table \ref*{table:Table_1}}. The results unequivocally demonstrate that the KDC-diff model, utilizing realistic-vision-v51 as its teacher, surpassed all competing models across the specified evaluation metrics. KDC-diff achieved the lowest FID score of 297.66, the lowest KID score of \(0.0017 \pm 0.0435\), and the lowest LPIPS value of 0.581. Furthermore, it recorded the highest CLIP score of 33.89, highlighting its superior capability to perform effectively across diverse datasets within a limited resources scenario.

 \hyperref[table:Table_2]{Table \ref*{table:Table_2}} presents the inference times and parameter counts for all evaluated models, including KDC-Diff variants on both datasets. Among the tested configurations, the combination involving the full-scale UNet (anything mid-journey) records the highest inference time of 13.0298 seconds and the largest parameter count of 859,520,964, along with a FLOPs score of 339.01 GMac on Oxford 102 Flower Dataset. In comparison, our proposed UNet variants demonstrate significantly improved efficiency. Specifically, KDC-Diff-x, KDC-Diff-y, and KDC-Diff-CL achieve inference times of 12.7688, 12.398, and 13.023 seconds, respectively, all with a reduced FLOPs score of 228.85. Notably, the final KDC-Diff model delivers the best overall performance, achieving the fastest inference time of 7.854 seconds, the lowest parameter count at 482,346,884, and a FLOPs score of 228.85—highlighting its lower complexity and superior computational efficiency. 
 
 Among the competing models tested on the Butterfly \& Moths dataset, realistic-vision-v51 emerged as the closest in performance, attaining an FID score of 302.58 and a CLIP score of 28.04. However, these metrics remain significantly inferior to the KDC-diff(Proposed Model). The efficiency of KDC-Diff is further highlighted in  \hyperref[table:Table_2]{Table \ref*{table:Table_2}}, where it achieves an average inference time of 7.94 seconds per image, a substantially lower parameter count of 482,346,884, and a FLOPs value of 228.85, which emphasizes its computational advantages.As depicted in \hyperref[fig:Figure_4]{Fig. \ref*{fig:Figure_4}}, the generated images visually reinforce the model's qualitative superiority and robustness in handling diverse and challenging datasets. The outputs demonstrate enhanced fidelity to the input prompts, improved structural consistency, and better semantic alignment compared to baseline models. These results suggest that our approach not only preserves key visual features across different domains but also generalizes effectively under varying levels of complexity and data distribution shifts.

Visualization of the attention map corresponding to the generated image has been illustrated in \hyperref[fig:Figure_5]{Fig. \ref*{fig:Figure_5}}. The grayscale heatmap highlights the spatial regions that the model focused on during the image generation process. Brighter areas in the map correspond to higher attention weights, indicating regions of greater significance in the model’s internal representation. These focused areas suggest where the model concentrated its learning capacity to encode key structural and semantic cues from the input prompt. By analyzing these attention distributions, we gain valuable insight into how the model dynamically allocates focus across different spatial locations, thereby contributing to the generation of coherent and contextually aligned outputs. This not only enhances the interpretability of the diffusion process but also reinforces the model's ability to preserve fine-grained details and maintain spatial consistency throughout the generative stages. Moreover, the visualization supports our claim that the proposed framework effectively captures and preserves salient regions, ensuring a better balance between global structure and local fidelity in the generated images.

\section{Discussion}
The proposed KDC-Diff framework establishes a new benchmark in efficient text-to-image generation by integrating architectural simplification with knowledge transfer and continual adaptation. At the core of our method lies a structurally compressed U-Net that reduces the model’s footprint without sacrificing expressive capacity. This is made possible by a dual-layered knowledge distillation mechanism that enables the student model to inherit both semantic understanding and intermediate representational richness from a larger, more expressive teacher model. The inclusion of latent-space CL further reinforces the model’s adaptability, allowing it to retain performance across sequential learning tasks and dynamic data distributions. Collectively, these contributions allow KDC-Diff to maintain high generative quality and semantic alignment while operating with a drastically reduced parameter count, computational cost, and memory usage. This unified approach demonstrates that performance and efficiency need not be mutually exclusive in the design of next-generation generative models.

When compared against a wide range of leading diffusion-based models—including Stable Diffusion v1.5, v2.1, Realistic Vision v5.1, and Anything-v5—KDC-Diff demonstrates consistent superiority across all primary evaluation criteria, including FID, CLIP, KID, and LPIPS. Beyond outperforming these models in image quality, semantic-text alignment, and perceptual consistency, KDC-Diff achieves these gains while operating at a fraction of their computational complexity. Whereas conventional baselines rely on architectures exceeding 850 million parameters, prolonged inference times, and high FLOPs costs, our proposed model delivers enhanced generative performance with significantly fewer parameters, faster inference, and markedly reduced computational overhead. These improvements remain consistent across challenging datasets such as Oxford-102-Flower and Butterfly \& Moth 100 Species, underscoring the model’s robustness and generalization capacity. Even when compared to high-performing architectures like Realistic Vision v5.1, KDC-Diff produces more coherent and semantically accurate outputs while maintaining lower architectural complexity. The model’s capacity to balance image fidelity, semantic alignment, computational efficiency, and adaptability positions it as a solid foundation to build efficient, generalizable, and semantically consistent generation pipelines tailored to low-resource environments.



\section{Conclusion}
In this study, we introduced KDC-Diff, a resource-efficient and high-performing framework for T2I generation, built upon Stable Diffusion. By integrating a structurally streamlined U-Net architecture with a dual-layered knowledge distillation strategy and a latent space replay-based CL mechanism, the proposed model significantly reduces computational overhead while preserving or even improving generative quality. The architectural modifications led to a substantial reduction in model size (482M parameters), inference time (7.85 seconds), and computational complexity (228.85 GMac FLOPs), making KDC-Diff highly suitable for deployment in constrained environments. Comprehensive evaluations on benchmark datasets, including Oxford-102-Flower and Butterfly \& Moth 100 Species, demonstrate that KDC-Diff consistently outperforms several state-of-the-art models in terms of FID, CLIP score, KID, and LPIPS while maintaining lower resource consumption. These results validate the effectiveness of our lightweight design and latent-aware training strategies. Looking ahead, this work lays the foundation for broader applications of diffusion models in settings with limited hardware resources, and future research could explore extending the framework to multi-modal tasks, real-time generation, or adaptive fine-tuning on edge devices to further enhance the accessibility and scalability of generative AI.

\section*{Declaration of competing interest}
We do not have any conflict of interest.






\end{document}